\documentclass[aps,pre,twocolumn,groupedaddress,showpacs,floatfix]{revtex4}
\usepackage{amsmath}
\usepackage{amssymb}
\usepackage{graphicx}

\begin{document}

\title{Complex Networks: New Concepts and Tools for Real-Time Imaging and Vision}

\author{Luciano da Fontoura Costa} 
\affiliation{Instituto de F\'{\i}sica de S\~ao Carlos. 
Universidade de S\~ ao Paulo, S\~{a}o Carlos, SP, PO Box 369,
13560-970, phone +55 16 3373 9858,FAX +55 16 3371 3616, Brazil,
luciano@if.sc.usp.br}

\date{20th Feb 2006}

\begin{abstract}
This article discusses how concepts and methods of complex networks
can be applied to real-time imaging and computer vision. After a brief
introduction of complex networks basic concepts, their use as means to
represent and characterize images, as well as for modeling visual
saliency, are briefly described.  The possibility to apply complex
networks in order to model and simulate the performance of parallel
and distributed computing systems for performance of visual methods
is also proposed.
\end{abstract}

\maketitle

\section{Introduction}
\label{sec:intro}

To a large extent, scientific advances have been constrained by the
increasing complexity of the data and dynamics under analysis.  Having
initially tamed the simplest systems by using linear systems or linear
approximations (and as a consequence missing the most interesting
dynamics), science and engineering progressed inexorably towards the
so-called \emph{complex systems} and structures.  Driven by
challenging applications in areas such as biology and neurosciences,
the developments in image processing and computer vision have been no
exception to this rule.  Along the last decades, not only has the
availability of high resolution images increased substantially, but
the problems under analysis have also became more and more intricate
as implied by the underlying non-linear dynamics and spatial
complexity.  For instance, advances in biotechnology and image
acquisition have allowed the capture of 4D images (3D plus time)
identifying gene expression in presence of anatomical structures
(e.g.~\cite{sectioning:2004}).  High complexity visual data has also
been continuously produced as results of simulations.  While the
capture and storage of such data implies their own technological
demands, the concepts and methodologies required for the analysis and
modeling of such results impose major challenges not only on the
theoretical side, but also on developing computational resources
capable to cope with the complexity of both data and algorithms.
The latter is particularly critical because of the recent slow down of
Moore's law.  In this text we briefly overview how the area of
\emph{Complex Networks}, by incorporating many flexible and powerful
concepts and tools, can provide the means for significant advances
in real-time imaging science and technology.

\section{Complex Networks}
~\label{sec:cns}

Complex networks and graph theory are closely intertwined areas.
However, while the latter concentrates on combinatorics and
mathematical theoretical formalism, the former emphasizes dynamics,
statistics, statistical physics, and applications.  Yet, the eventual
differences between these two areas should be considered for their
potential to complement one another, and not as a divisive barrier.
While random networks have been around for a long time
(e.g.~\cite{Bollobas:2004}), their applicability has been constrained
by the fact that few structures and phenomena in nature seem to follow
uniformly random organization~\footnote{In a random network, the
probability of connection between any pair of nodes is constant.}.  It
was mainly through the studies of small world properties of social
networks (e.g.~\cite{Watts:2004}), jointly with the more recent
characterization of \emph{power law} (or \emph{scale free}) networks
(e.g.~\cite{Albert_Barab:2002}), that the area of complex networks
gained momentum to become one of the most exciting modern fields of
research.  The success of complex networks stems from several factors,
especially: (i) the generality of networks (and graphs) to model
virtually any discrete structure; (ii) the fact that natural
structures often present small world or scale free connectivity; and
(iii) the possibility to relate the network topology and dynamics.
Although relatively young, the applications of complex networks
already extend from genetics to scientific citations.  Complex
networks have been reviewed in comprehensive surveys
(e.g.~\cite{Albert_Barab:2002,Newman:2003,Boccaletti_etal:2006,Costa_surv:2005})
and books (e.g.~\cite{Linked,Watts:2004,Dorog:2004}).

Of particular interest to imaging and vision are the special category
of networks known by the name of ~\emph{spatial} or
~\emph{geographical}.  Such networks are characterized by the fact
that each constituent node has a specific, well-defined position
within an embedding space.  Examples of spatial networks include
highways, air routes between airports, power distribution systems,
among many others.  Network representation of images are inherently
spatial, as the position of pixels, regions or objects associated to
the networks will have well-defined coordinates.  A particularly
interesting possibility is to consider spatial networks where the
positions of the nodes vary along time, as is the case with mobile
communications or video.  In addition to their connectivity, spatial
networks are also characterized by the spatial distribution of the
nodes along space.  Because of the adjacency imposed by the most
immediate neighbors, spatial networks tend not to exhibit the small
world property, unless long range connections are incorporated.  At
the same time, the strength of the interconnection between pairs of
nodes usually tend to decrease with the distance between those points.
A particularly interesting concept for analysis of the spatial
structure of such networks is the Voronoi diagram
(e.g.~\cite{Costa_poly:2006}).

\section{Complex Networks for Image Representation}

Image analysis and recognition involve the integration of information
and clues from local properties of the pixels (e.g. gray-level, color)
to neighborhoods, objects and scenes (e.g.~\cite{Costa_Shape:2001}).
In other words, the integration of the visual information takes place
along several scales of space and time, from micro to macro, and
vice-versa (e.g.~\cite{Tovee:1996}).  While many shape and image
analysis methods have relied on the use of pixels as basic elements
(e.g.~\cite{Costa_Shape:2001}), the representation of an image as a
complex network allows the short, medium and long range relationships
between pixels as objects to be explicitly represented and taken into
account.  Therefore, it would be interesting to conceive algorithms
taking into account not only the raw image, but also its enhanced
representation as a complex network.

There are several possibilities for transforming an image into a
complex network.  In this article we restrict our attention to the two
following methodologies.  In the first, each pixel is associated to a
node, while edges are defined in terms of the similarities between the
local properties of pixels~\cite{Costa_vision:2004,Costa_hub:2004}.
For instance, it is possible to associate every pair of pixels through
an edge whose weight is inversely proportional to the difference
between the respective gray-levels.  Other local features such as
gradient magnitude and orientation, local dispersion, and color, can
also be considered for defining the similarities and weights.  In
order to simplify the otherwise densely networks obtained by such a
methodology, it is often useful to establish a threshold for
connections (i.e. only the most similar pixels are connected) and to
consider for connections only those pixels lying within a fixed
distance from the reference node
(e.g.~\cite{Costa_bioinfo:2006,SPIE}).  The second methodology for
converting an image into a graph involves the determination of tangent
(or normal) fields along the image contours and interconnecting all
pixels falling under the straight lines determined by such
fields~\cite{Costa_sali:2006} (more detail in Section~\ref{sec:sali}).

\section{Complex Networks for Image Characterization and Analysis}

Once an image is represented as a complex network, it becomes possible
to devise more comprehensive algorithms for its characterization and
analysis which rely not only on local properties (i.e. pixels and
respective neighborhoods), but also on medium to long range
relationships.  For instance, the node degree~\footnote{The degree of
a network node is equal to the number of edges connected to that
node.}, as well as many other measurements associated to individual
nodes (e.g. clustering coefficient~\cite{Albert_Barab:2002} and
hierarchical
measurements~\cite{Costa_prl:2004,Costa_hier:2005,Costa_gen:2006}) can
be used as subsidy for identification of
textures~\cite{Costa_vision:2004,SPIE} and even objects.  Another
example of a typical image analysis situation where complex networks
presents promising performance is \emph{image segmentation}, i.e. the
task of identifying the parts of the image corresponding to the
existing objects in the image.  Of particular interest is the
application of \emph{community finding} algorithms.  Loosely speaking,
a \emph{community} (e.g.~\cite{Newman_comms:2006}) is a portion of the
network (a subnetwork) whose nodes are more intensely related one
another than with the rest of the network.  The problem of identifying
communities in networks is challenging and closely related to problems
in pattern recognition~\cite{Costa_Shape:2001}, sharing methodologies
and limitations.  Such algorithms can be applied to images represented
as networks so that each community will, in principle, correspond to a
segmented region~\cite{Costa_hub:2004}.  Because the network
representation typically supplies more information than just the
pixels and their immediate neighborhood, it is expected that such
algorithms may lead to enhanced performance when compared to
traditional image segmentation methodologies.

\section{Complex Networks for Visual Saliency Detection}
\label{sec:sali}
In this section we illustrate the possibility of integrating the
topological and dynamical properties of an image represented as a
complex network~\cite{Costa_sali:2006}.  First, the image is
transformed into a complex network by representing the edge elements
in the images as nodes and defining the network links as follows.  For
each high contrast pixels $p$ (i.e. an edge) of the image, its
orientation $\alpha$ (e.g. direction of the gradient at $p$) is
estimated and used to define a straight line $L$ throughout the image
which passes through $p$ and has orientation $\alpha$.  All other edge
elements of the image which fall under the line $L$ are connected
bidirectionally to $p$.  Now, a random walk is performed on the
obtained network which proceeds as follows: being currently at node
$p$, the next movement is determined at random between the edges
emanating from $p$.  The steady state solution of such a dynamics can
be conveniently obtained in terms of the eigenvector equation
$\vec{q}=W \vec{q}$, where $W$ is the stochastic matrix associated to
the Markov chain driven by the random walk and $\vec{q}$ is the state
occupancy at equilibrium.  It has been shown~\cite{Costa_sali:2006}
that the nodes with highest values of occupancy ratio tend to
correspond to points of high saliency in the image.  Actually, such a
basic framework can be immediately extended to treat other situations
in saliency detection such as those involving elementary, pre-defined,
saliency indices associated to each object or
point~\cite{Costa_sali:2006}.  This example also illustrates how
complex networks can be defined with nodes associated not to pixels,
but to pre-defined regions or objects at larger spatial scales.

\section{Complex Networks for Modeling and Quantification of Performance}

Another area where complex networks can provide valuable help for
those working on real-time image processing and analysis concerns the
characterization, modeling and simulation of the performance of
real-time distributed systems.  Distributed systems can be immediately
mapped as complex networks -- each processor as a node, with
respective interconnections -- and the effect of different
interconnections models (e.g. random, small world, scale free, regular
lattices) over the respective performance quantified through
simulations.  Such a possibility has been preliminary addressed
regarding the performance of \emph{grid} computing
systems~\cite{Costa_grid:2005}, where the performance while
distributing computer tasks was quantified in terms of speed-up
measurements.  We are currently applying a similar methodology for
investigating the performance of distributed systems interconnected
through different network models while processing a stream of
real-time video frames.

\section{Concluding Remarks}
~\label{sec:crs}

This article has overviewed the perspectives of the new area of
complex networks for coping with the demands being imposed onto
imaging and real-time processing as a consequence of ever increasing
complex data and algorithms.  It has been argued that, because of
their generality for representation of discrete structures and
emphasis on statistics and dynamics, complex networks can greatly
contribute to advances in areas including image characterization and
segmentation as well as the analysis of performance of real-time
parallel and distributed systems.  Illustrations of the potential of
such a methodology have been provided with respect to several typical
problems in real-time image processing and analysis, including image
segmentation, texture characterization, saliency detection at varying
spatial scales, and quantification and modeling of distributed systems
for video processing.  Because of the incipiency of such applications,
there is plenty of room for related developments and investigations.

\begin{acknowledgements}
Luciano da F. Costa thanks CNPq (308231/03-1) for sponsorship.  
\end{acknowledgements}

\bibliographystyle{apsrev}
\bibliography{arxive}   

\end{document}